# Robot's Inner Speech Effects on Trust and Anthropomorphic Cues in Human-Robot Cooperation*

A. Pipitone, A. Geraci, A. D'Amico, V. Seidita, and A. Chella

*Abstract—* Inner Speech is an essential but also elusive human psychological process which refers to an everyday covert internal conversation with oneself. We argue that programming a robot with an overt self-talk system, which simulates human inner speech, might enhance human trust by improving robot transparency and anthropomorphism. For this reasons, this work aims to investigate if robot's inner speech, here intended as overt self-talk, affects human trust and anthropomorphism when human and robot cooperate. A group of participants was engaged in collaboration with the robot. During cooperation, the robot talks to itself. To evaluate if the robot's inner speech influences human trust, two questionnaires were administered to each participant before (pre-test) and after (post-test) the cooperative session with the robot. Preliminary results evidenced differences between the answers of participants in the pre-test and post-test assessment, suggesting that robot's inner speech influences human trust. Indeed, participant's levels of trust and perception of robot anthropomorphic features increase after the experimental interaction with the robot.

## I. INTRODUCTION

Trust is multifaceted psychological construct with no universal definition, whose research field has fed upon contribution from many various disciplines. From a psychological viewpoint, there are two main perspectives on human-human trust: on one hand, trust is conceived as a stable trait, shaped by human early life trust-related experiences, which highlight a dispositional propensity to trust others [1,2]. On the other, trust is described as a changing state affected by cognitive, emotional and social processes [3,4]. More broadly, scholars agree that trust involve two main features: one's positive attitudes and expectations on the trustee [5] and one's willingness to being vulnerable and to accept risks [6].

In the past years, trust have become one of the leading research topic in the field of human-machine interaction, since artificial systems development and implementation have increased exponentially in every context, leading to growing interactions with humans [7]. In particular, robots are now used in different contexts such as military, security, medical, domestic, and entertainment [8]. Robots, compared to other automations, are designed to be self-governed to some extent, in order to respond to situations that were not pre-arranged [9]. Therefore, the greater the complexity of robots the higher the importance of trust in human-robot interaction [10].

### A. Trust in HRI

In human-robot interaction (HRI) literature, trust is considered to be a key factor in human reliance on robot partner [9, 10] since it is defined as an «attitude that an agent will help achieve an individual's goals in a situation characterized by uncertainty and vulnerability» [10]. Recently, HRI studies supported a three-factor model that blends the empirical contributions gathered so far in human-robot trust [11, 12]. According to this model, human-robot trust dynamically emerges from the interaction among human-related factors (e.g. personality traits, emotional and cognitive processes), robot-related factors (e.g. intelligence, transparency, anthropomorphism) and environment-related factors (e.g. competitive/collaborative context, culture, physical environment) [11, 12]. Regarding the robot-related factors, studies have shown that people tend to trust more those robots that look (i.e. head, body, face, voice) and behave (e.g. non-verbal elements, dyadic and social gestures) like humans [13, 14]. In addition, trust is enhanced when people have a clear understanding of why, when and how a robot operates [15, 16], that's because a system transparency help humans to form a precise mental model of robot capabilities [16]. It is crucial for humans to have a clear understanding how and why a robot works, considering that trust may be at risk when robot capabilities cannot be comprehended [17]. As a consequence, new automation system should be designed considering such evidences from empirical research in order to facilitate human-robot collaboration.

### B. The Role of Inner Speech in Human-Robot Trust

In psychological literature, inner speech is a well-known construct that was first theorized by Vygotsky who conceived it as the result of a set of developmental processes [18]. Continuous linguistic and social interaction between the child and the caregiver are progressively internalized and take the form of covert self-directed speech. In time, the child gradually becomes more autonomous and gain the ability of self-regulation. Scholars have used different terms when referring to inner speech (e.g. covert speech, self-talk, private speech). However, it is generally defined as the subjective experience

*Research supported by Air Force Office of Scientific Research under award number FA9550-19-1-7025.

A.P. First Author is with Engineering Department, University of Palermo, Viale delle Scienze, Building 6, IT (corresponding author e-mail: arianna.pipitone@unipa.it).

A.G. Author is with Psychology Department, University of Palermo, IT (e-mail: alessandro.geraci@unipa.it).

A.D. Author is with Psychology Department, University of Palermo, IT (e-mail: antonella.damico@unipa.it).

V.S. Author is with Engineering Department, University of Palermo, IT (e-mail: valeria.seidita@unipa.it).

A.C. Author is with the Engineering Department, University of Palermo, Building 6, IT, and associated with the National Research Council (ICAR-CNR), Palermo, IT (e-mail: antonio.chella@ unipa.it).

of language in the absence of an audible articulation [19]. There are some evidences that inner speech plays an important role for human psychological balance as it is linked to self-awareness [20], self-regulation [21], problem-solving [22], and adaptive functioning [19]. Recently, innovative computational model has been developed which pave the way to new frontiers in the field of artificial intelligence. In particular, it has been proposed a cognitive architecture for implementing inner speech in robot [23]. More specifically, since inner speech is a covert speech that cannot be heard from the outside, robot's inner speech is reproduced using overt self-talk. The same architecture is now used in the field of human-robot interaction, considering that robot's inner speech might improve robot transparency and anthropomorphism, ultimately enhancing human-robot trust [24]. Moreover, the same architecture was used for demonstrating how robot inner speech improves the robustness and the transparency during cooperation, meeting the standard requirements for collaborative robots [25]. Suggestive results were also obtained in passing the mirror test: inner speech enables a conceptual reasoning for inferring the identity of the reflected entity in a mirror, and robot becomes able to recognize itself [26]. We argue that robot's inner speech might act as facilitators for human understanding and predicting the robot behaviors, as they form adequate mental representation of the robot. In addition, such system, which simulates a human psychological functioning, would facilitate user attribution of human qualities to the robot ultimately improving human-robot trust. Taking all this into account, our study aims to investigate if the interaction with a robot equipped with inner speech system improves trust levels and perception of robot features in terms of anthropomorphism, animacy, likeability intelligence and safety, during the execution of a cooperative task. In addition, this study aims to investigate if participants use of inner speech in everyday life moderates the effects of robot's inner speech on trust levels and perceptions of robot in human-robot collaboration:

$H^1$: Mean level of trust in the experimental group is higher than the control group;

$H^2$: Mean level of trust in the experimental group is higher in the post-test than the pre-test;

$H^3$: Mean level of perception of robot feature in the experimental group is higher than the control group;

$H^4$: Mean level of perception of robot feature in the experimental group is higher in the post-test than the pre-test;

$H^5$: Mean level of trust tend to increase when participants use of everyday self-talk is high.

## II. METHOD

To analyze the effects of robot's inner speech in trust when humans and robot collaborate, a simple scenario was defined in which they have to cooperate for achieving a common goal.

The scenario consisted to virtually set up a table with the robot, following an *etiquette schema*. The schema defines the set of rules according to which the utensils have to be arranged in the table. Fig. 1 shows the etiquette schema used in the experiments. If an utensil is finally placed on a different position than the expected one according to the schema, the etiquette rule for that utensil is infringed.

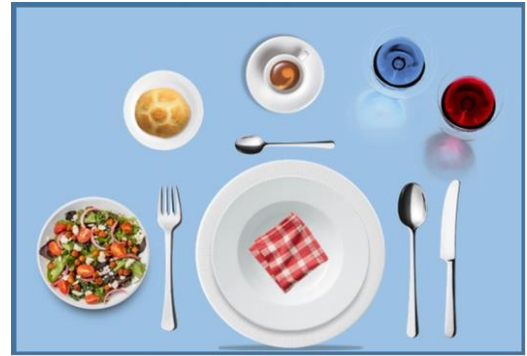

*Fig 1. The etiquette schema defining the rules for setting up the table*

The virtual table is implemented on a tablet surface, where the participant can drag and drop the utensils, can make requests to the robot, and can see the robot's actions. The choice of that scenario enabled the possibility to analyze the cues in particular situations which occur during human-robot cooperation, that are:

- the *etiquette infringement*, representing a conflictual situation, that is the participant places the utensils in an incorrect final position, or he/she asks to the robot to place an object in a position which infringes the etiquette; the conflict arises because the action is not allowed, and the human and the robot have to decide how to continue. In some cases, the human can decide to infringe the rule, or to repeat the action to be compliant with the schema.

- the *discrepancy situation*, that is the participant asks the robot to pick an object already on the table.

When human and robot cooperate to set up the table, an important aspect regarded the definition of the kind of the dialogue the robot implements, including inner and outer turns. The linguistic form of the sentences in the turns were differentiated for inner and outer speech in order to evaluate the impact of the inner speech when it is activated in the experimental session compared to the control session in which inner speech is not activated. It allows to analyze the impact of robot's inner speech in the cues in human-robot interaction. The subsection A details the dialogue properties and the experimental setup definition.

Another aspect regarded the implementation of the virtual environment. The scenario of a table in which to place utensils according to the etiquette rules, was simulated by an Android app running on a 15" tablet. The app was integrated with the typical robot routines for enabling the robot to the event detection in the virtual table, and to the virtual action execution. The requests to the robot were simulated by a list of checkbox, and the participant can choose one of them at a time. In this way, each participant can do the same kind of questions, enabling the same observations on the whole participants. All these implementation features are detailed in the subsection B.

| Table I. Differences between Robot Outer Speech and Inner Speech | |
|---|---|
| **Outer Speech** | **Inner Speech** |
| Always produced | At times produced |
| Experimental and control group | Experimental group |
| Short sentences | Short/medium sentences |
| Objective feedback | Personal statements, comments |
| Formal language | Informal language |

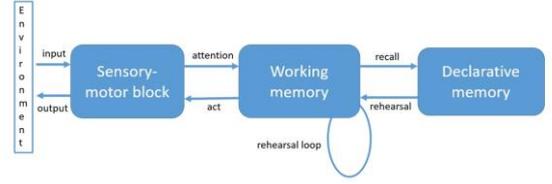

*Fig. 2. The outline of the cognitive architecture of inner speech*

### A. Research Design

For this study we plan a pre-test/post-test control group design. Participants were divided in two different groups, one experimental group and one control group. In the experimental group, participants will interact with the robot equipped with inner speech whereas in the control group participants will interact with the robot that produces only outer speech. Both groups will be administered with research protocols which include psychological measures (see section E) in order to detect differences between experimental and control groups and also between pre-test and post-test sessions. However, due to the outbreak of COVID-19 the study has slowed down, so to this date, participants for the control group have yet to be recruited. Consequently, this paper presents only preliminary results obtained in the experimental group

### B. Kinds of Inner Speech and Dialogue

In order to present the same stimuli in both experimental and control groups the structure of robot outer and inner speech was defined prior to the experiments (Table I). Participants can set up the table either moving objects on their own or asking the robot to do it.

Either way, the robot will produce a vocal response in the form of outer speech followed by the inner speech only in the experimental condition. Outer speech follows the typical language that is expected by an artificial agent, as it uses formal language and it only gives objective feedback based on the participant's performance and actions. On the contrary, inner speech traces a human-based language, since it expresses robot values, personal statements and comments on participant's performance and actions using a friendly and colloquial form.

### C. Implementing Inner Speech Skill in Robot

The robot's inner speech is implemented by the cognitive architecture proposed by some of the authors [23]. An outline of the architecture is shown in Fig. 2. The core of the architecture is the working memory: it decodes input signals from the environment, perceived by the sensory-motor block, and associates to them symbolic information (labels). Generally, this process is the output of typical routines, as speech-to-text routines which decode audio in sequences of words, or neural networks which extract the content of an image and associates to each recognized entity the corresponding word.

The working memory recalls from the declarative memory concepts related to the labels. The declarative memory represents the domain knowledge. By recalling related concepts, new words emerge and they are in turn decoded by the working memory, as they were perceived from the environment. They are processed as the labels, and a rehearsal loop starts. The inner speech is this rehearsal loop enabling the emergence of another concepts and topics in the working memory. By inner speech, a form of reasoning starts, and in some case, it could be necessary to take actions on the environment (for example, it could be necessary to perceive new information from the environment, or to move an object, and so on). These actions are executed by the same sensory-motor block.

In the proposed scenario, the inner speech is a bit differently implemented within the cognitive architecture, with the aim to enable the observations of the specific cues. In particular, to analyze the cues in the same conditions for each participant, the inner and outer dialogue of the robot has to involve the same turns for the same events. In this way, the participants' evaluations about the interaction depend on the same variables and parameters, and the evaluations can be compared for abstracting a general inner speech affection on the interaction.

For this reason, the inner speech cognitive architecture functioning was simplified in respect to the aforementioned completed version. The main differences regarded the decoding of the perception and the emergence of the semantic content of the dialogue. In the experiments, the environment is virtual and the perception just regarded the actions the participant does on the tablet surface. To each action executed by the participant corresponds an event that is detected by the robot (the robot *perceives* the event). The event can involve a wrong or a correct action in respect to the etiquette rules, a request to the robot to do something, and so on. According to the cognitive architecture, the event is decoded by the working memory. While, in the original version, the working memory decodes environmental signals by associating to them labels, now the working memory associates to each event a numerical symbol, which univocally identified that event.

To that symbol corresponds a sentence in the declarative memory (in this case, it works as a vocabulary of sentence, by returning the sentence corresponding to a symbol). Just the corresponding sentence to the specific event is recalled from the declarative memory. The sentence is then produced and re-hear, recalling from the declarative memory the next turn of the dialogue. The involved turns could be inner or outer sentences, that are produced according to a specific protocol, as detailed in section II. That protocol aims to define typical turns in the interactions, according to the participant's expectation. For example, the participant will wait always for a vocal feedback by the robot, so the robot will always produce one or more outer sentences. Instead, the participant does not always attend inner speech, and the inner dialogue is not always produced by the robot. Obviously, the involved sentences have a specific meaning that are semantically related to the event or to the previous re-hear sentence. They are recalled from the declarative memory according to the previous aforementioned order, so a disambiguation strategy was not necessary.

For example, let suppose the participant (named Bill) asks the robot to place the knife in a wrong location on the table, that is to the left of the plate, while it has to stay to the right. In this case, the event is a request to the robot to infringe the etiquette. The robot perceives that event, and the working memory associates the numerical identifier to it. It recalls from the declarative memory the first sentence of the dialogue, and the loop starts, by recalling the other sentences, that are in turn (I stays for inner sentence, O for outer sentence):

I: "To make this request, Bill does not know that the knife should not be placed in that position or he wants to test me."

I: "Should I put the knife to the left of the plate? But if it goes right! "

O: "Bill, do you really want to infringe the etiquette rule for the knife?"

CASE 1: Bill answers yes

Bill: "yes, I do!"

I: "I don't want to disappoint him…"

O: "Ok Bill, I will place the knife to the left of the plate, as you want."

CASE 2: Bill answers no

Bill: "No!"

O: "Great! I will place the knife in the position expected for it!"

I: "I must pay attention; the knife is dangerous!"

I: "But I'm robot, the knife never hurts me"

O: "Knife moved to the right of the plate!"

The participant listens all the turns, that are produced by setting different parameters for inner and outer sentences. In this way, the participant becomes able to discriminate the self-talk to the dialogue with him/herself, and can evaluate the potential of inner speech during interaction.

In particular, the parameters involve the tune and the volume of the voice, the color of the leds of the robot and the double effect in the voice, that is activated while producing inner sentence, for giving a mentalizing effect of the voice.

*D. The Platform for Virtually Setting Up the Table*

The virtual environment for setting a table was implemented by an Android app, designed and built by the Mit App Inventor platform by the Massachusetts Institute of Technology[1]. Specifically, the virtual environment for the experimental session regarded the following issues:

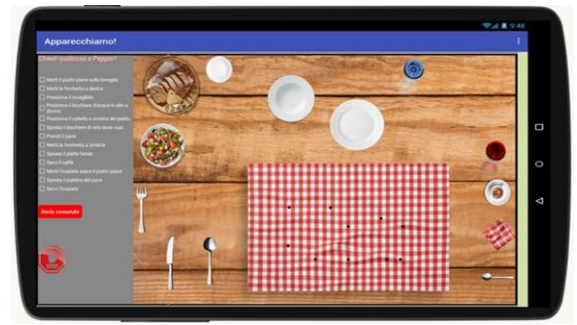

*Fig. 3: The app interface for cooperating with the robot by the tablet*

- the *app design and building,* for running it on the tablet used by the participants;
- the *event detection* strategy, for capturing the actions executed by the participant by the tablet, that is the evaluation of the final location in which he/she places the utensils, or the request he/she makes to the robot by the checkbox list;
- the *action execution* strategy, for allowing the robot to place utensils on the tablet according to the participant's request or based on its autonomous choices.

Fig. 3 shows the app interface, which looks very intuitive. The interface includes a main canvas with the table and utensils representation, and a lateral bar containing the list of checkboxes for the requests to the robot. Moreover, the lateral bar includes the stop button for ensuring the participant to stop at any time he/she wants. At the start of the experimental session, the utensils to locate are sparse on the table, and they have to be placed on the table cloth according to the etiquette rules.

The table cloth was marked by black dots, for highlighting the possible correct final locations. In this way, the participant has just the burden to select which objects to place in which dot, reducing the degrees of freedom.

The communication between the robot and the app was implemented by a hybrid client-server architecture. Fig. 4 shows the whole platform. The *central node*, represented by a computer, handles synchronous network requests. The node is hybrid because it runs as client or server according to the item with which it interfaces. In particular, the node will be:

- the *client*, when it requests to the robot to do something (to speech, to execute a virtual action, to track the participant, and so on). In this case, the server is the *proxy* of the robot, implemented by the Aldebran library (ALProxy) for *naoqi* developer[2], which switches the client's request to the typical robot's services (Speech, Track, Leds, and so on);

---

[1] https://appinventor.mit.edu/

[2] http://doc.aldebaran.com/2-5/index/dev/guide.html

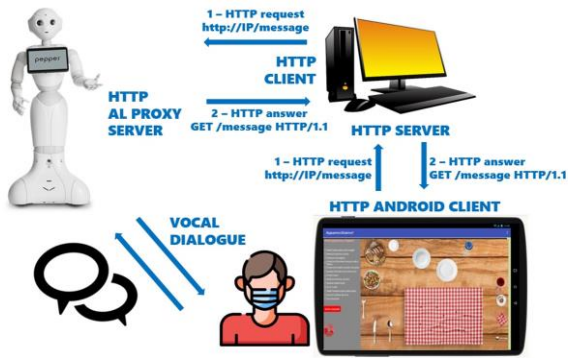

*Fig.4: The platform for making communication between the app and the robot*

- the *server*, when it receives request by the app, that will be in turn switched to the robot's proxy.

The robot-app communication involves the following use cases with corresponding kinds of requests:

- *the robot has to execute a virtual action*: when the participant selects a command in the lateral bar and clicks the Send Command button, the robot should to execute the specific action (it should to move an utensil on the tablet). In this case, the app sends to the node the request specifying the action to take, and the node forwards it to the robot. The request to the proxy will involve the aforementioned service, and the robot could dialogue with itself, or with the participant, or execute the action by answering to the node.
- *the participant executes an action*: when the participant drags and drops an utensil on the tablet screen, and finally he/she touches up the utensil, the final position could be on a correct dot, or not. The app detects such an event and sends to the node the information of correct or incorrect final location. The node forwards the message to the robot's proxy, and it calls one of the aforementioned services.

During the interaction, the robot can decide to do something (for example, it might refuse to execute a participant's request). The interaction ends almost 15 minutes after it starts, and the robot will autonomously decide to stop the session by informing the participant.

### E. Materials and Procedure

27 volunteers have been recruited using social network and they have been presented with the informed consent and COVID-19 protocol. Questionnaires have been administered to the participants through online form both in pre-test (Research Protocol A) and post-test (Research Protocol B). Research Protocol B has been administered after 15 days from Research Protocol A. The interaction session took place in the Robotics Lab where Anti-COVID measures have been addressed.

Data collection for the psychological variables included the following instruments:

- *Trust Perception Scale-HRI* [27] that assesses human's perception of trust in robots. The shortened version of the scale was used to make it easier for participants to fill out the questionnaires. The scale is composed of 15 item which provide a total score of trust.
- *The GOSDSPEED Questionnaire* [28] that assesses human's perceptions and impressions of a robot. It is one of the most used measurement tool to assess perceptions of robot [29]. It is a 24 item rating scale, that consists of a set of bipolar pair of objectives rated on a 5-point scale. The scale measures five different robot features: Anthropomorphism (5 items), Animacy (6 items), Likeability (5 items), Perceived Intelligence (5 items), and Perceived Safety (3 items)-
- *Self-Talk Scale* [30] that measures how frequently participants use inner speech in everyday life. It consists of 16 items scored on a 5-point Likert-type scale (from 0 = Never, to 4 = Very Often). The scale also measures four different dimensions of inner speech from 4 item each: Self-Criticism, Self-Reinforcement, Self-Management, Social Assessment.

### F. Preliminary Results

Due to the absence of control group data, these results were carried out considering only pre-test and post-test data from the experimental group.

In order to test hypothesis 2: "*Mean level of trust in the experimental group is higher in the post-test than the pre-test*" and hypothesis 4: "*Mean level of perception of robot feature in the experimental group is higher in the post-test than the pre-test*" we carried out a series of paired sample t-test for examining the differences in pre-test and post-test for each variable (Table II). Preliminary results confirm our hypotheses showing that within the experimental group, post-test mean scores in all the study variables are significantly higher than the pre-test mean scores, except for the dimension of the perceived safety to which the research hypothesis is rejected. These results seem to suggest that after the interaction with a robot equipped with inner speech, participants trust and perceptions of anthropomorphic features tend to increase. In addition, a moderation hypothesis was tested in order to investigate if such differences may vary depending on participants use of inner speech in everyday life. Thus, a within subject ANOVA test was carried out, entering self-talk scale mean scores as covariate. However ANOVA results were not significant, suggesting that participants use of everyday inner speech does not affect trust variation in pre-test and post-test session.

Table II.. T-Test results for mean differences between pre-test and post-test sessions.

| Variable | N | Pre-Test | | Post-Test | | $T^{(gl=26)}$ |
|---|---|---|---|---|---|---|
| | | M | SD | M | SD | |
| Trust | 27 | 65.70 | 7.60 | 68.91 | 6.91 | -2.06* |
| Anthropomorphism | 27 | 2.70 | .68 | 3.31 | .66 | -4.20*** |
| Animacy | 27 | 3.19 | .61 | 3.78 | .51 | -4.43*** |
| Likeability | 27 | 4.10 | .56 | 4.30 | .65 | -1.40* |
| Perceived Intelligence | 27 | 3.90 | .66 | 4.19 | .59 | -2.47* |
| Perceived Safety | 27 | 4.07 | .59 | 4.04 | .58 | .26 |

* p < .05, *** p < .001

## III. CONCLUSIONS AND FUTURE WORK

Preliminary analyses highlight some promising results showing that participants' trust levels and perception of robot anthropomorphic features increase in between pre-test and post-test session. Nevertheless, to this date, there are two main limitations that affect these results: the lack of control group data and the small sample size. The lack of a control group, in particular, does not allow to disentangle if such increasing scores in trust and perception of robot's anthropomorphic features depend whether on the presence of inner speech or simply on the interaction with the robot. However, these preliminary results appear to be promising and they could open new research frontiers. This study is still in progress so more data is yet to be gathered.